\documentclass[10pt,twocolumn,letterpaper]{article}

\usepackage[pagenumbers]{iccv} 
\definecolor{iccvblue}{rgb}{0.21,0.49,0.74}
\usepackage[pagebackref,breaklinks,colorlinks,allcolors=iccvblue]{hyperref}

\usepackage{textcomp}

\def\paperID{*****} 
\def\confName{ICCV}
\def\confYear{2025}

\title{HOSt3R: Keypoint-free Hand-Object 3D Reconstruction from RGB images}

\author{
  Anilkumar Swamy$^{1,2}$ \hfill
  Vincent Leroy$^{1}$ \hfill
  Philippe Weinzaepfel$^{1}$ \hfill
  \\[0.1cm]
  Jean-Sébastien Franco$^{2}$ \hspace{1.0cm}
  Grégory Rogez$^{1}$ \\[0.4cm]
  $^1$NAVER LABS Europe \hspace{0.2cm}
  $^2$Inria centre at the University Grenoble Alpes
}

\usepackage[capitalize]{cleveref}
\crefname{section}{Sec.}{Secs.}
\Crefname{table}{Table}{Tables}
\crefname{table}{Tab.}{Tabs.}   
\usepackage{multirow}
\usepackage{float}

\newcommand{\briac}[0]{FDR\xspace}
\newcommand{\sigg}[0]{HHOR\xspace}
\newcommand{\vh}[0]{VH\xspace}

\begin{document}
\maketitle

\begin{abstract}
Hand-object 3D reconstruction has become increasingly important for applications in human-robot interaction and immersive AR/VR experiences. A common approach for object-agnostic hand-object reconstruction from RGB sequences involves a two-stage pipeline: hand-object 3D tracking followed by multi-view 3D reconstruction. However, existing methods rely on keypoint detection techniques, such as Structure from Motion (SfM) and hand-keypoint optimization, which struggle with diverse object geometries, weak textures, and mutual hand-object occlusions, limiting scalability and generalization. As a key enabler to generic and seamless, non-intrusive applicability, we propose in this work a robust, keypoint detector-free approach to estimating hand-object 3D transformations from monocular motion video/images. We further integrate this with a multi-view reconstruction pipeline to accurately recover hand-object 3D shape. Our method, named HOSt3R, is unconstrained, does not rely on pre-scanned object templates or camera intrinsics, and reaches state-of-the-art performance for the tasks of object-agnostic hand-object 3D transformation and shape estimation on the SHOWMe benchmark. We also experiment on sequences from the HO3D dataset, demonstrating generalization to unseen object categories.
\end{abstract}    
\section{Introduction}
\label{sec:intro}

\begin{figure}[t]
    \centering

    \begin{minipage}{\columnwidth}
        \centering
        \begin{tabular}{@{}cccc@{}}
            \makebox[0.22\columnwidth]{\centering \textbf{Input RGB}} &
            \makebox[0.22\columnwidth]{\centering \textbf{View 1}} &
            \makebox[0.22\columnwidth]{\centering \textbf{View 2}} &
            \makebox[0.22\columnwidth]{\centering \textbf{View 3}}
        \end{tabular}
    \end{minipage}
    
    \vspace{1pt}

    \includegraphics[width=\linewidth]{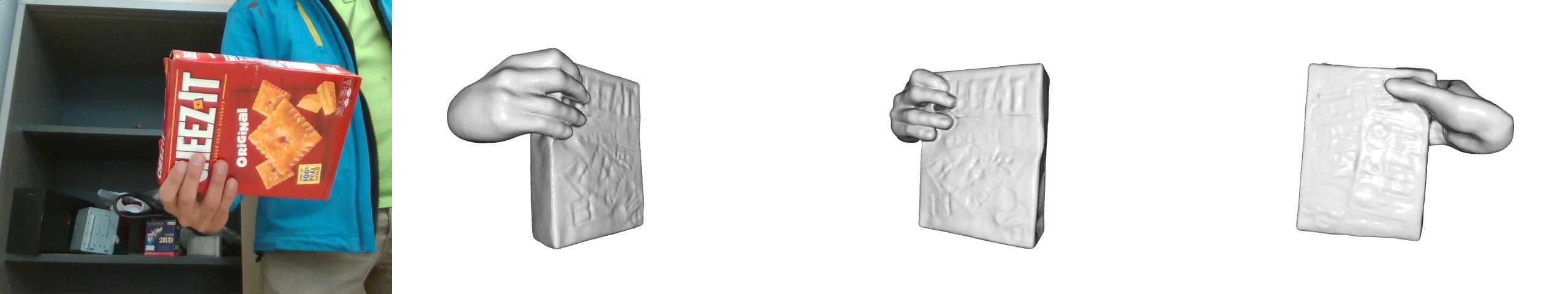}
    \includegraphics[width=\linewidth]{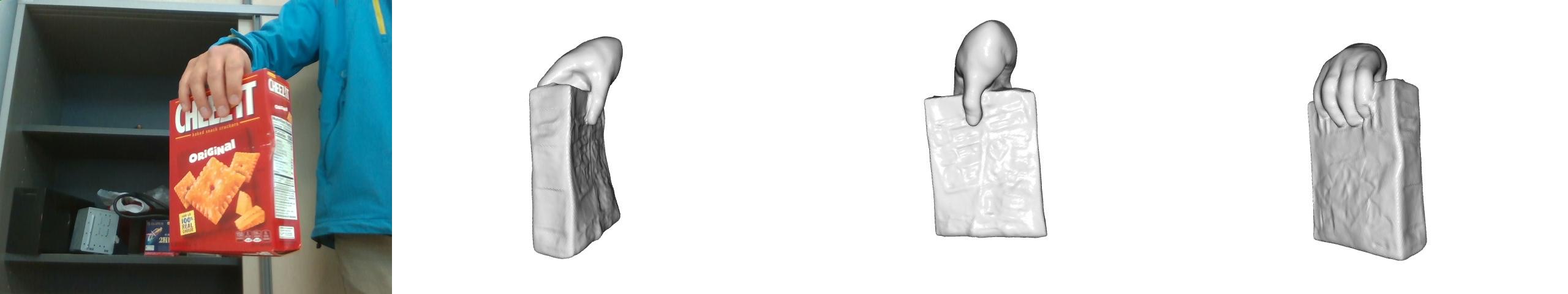}
    \includegraphics[width=\linewidth]{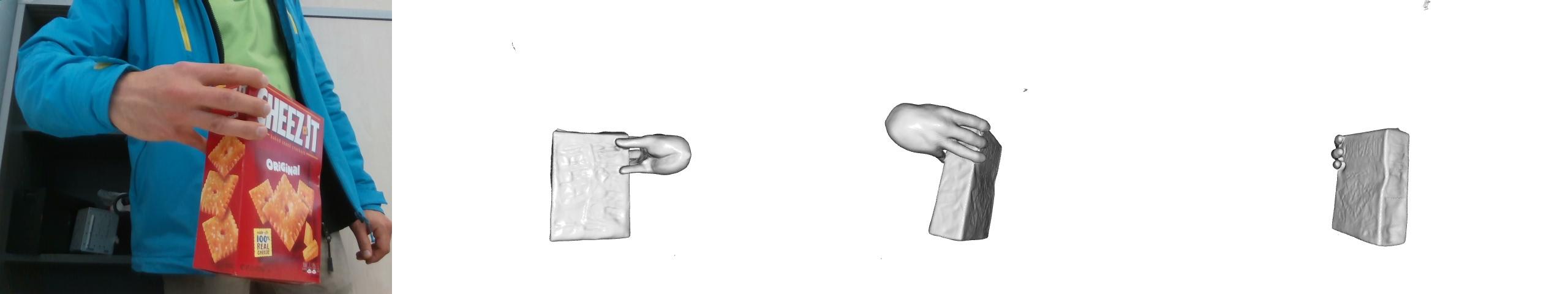}
    
    \includegraphics[width=\linewidth]{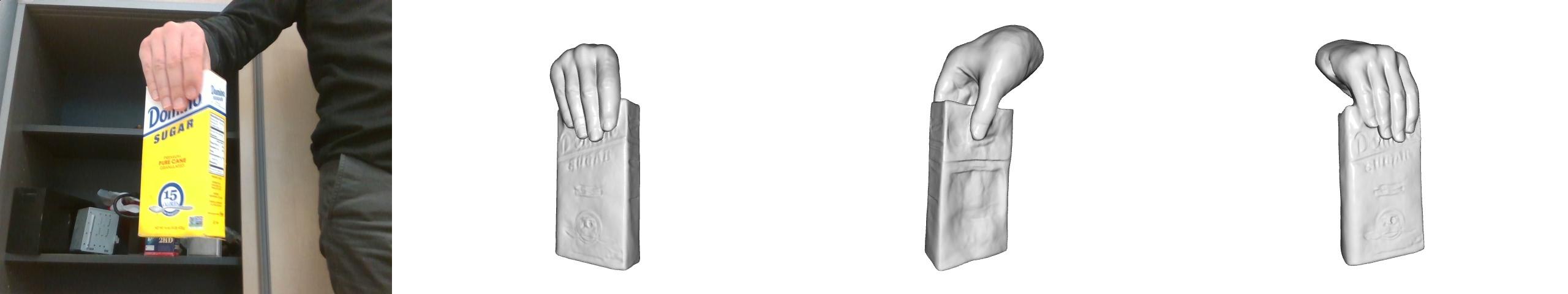}
    \includegraphics[width=\linewidth]{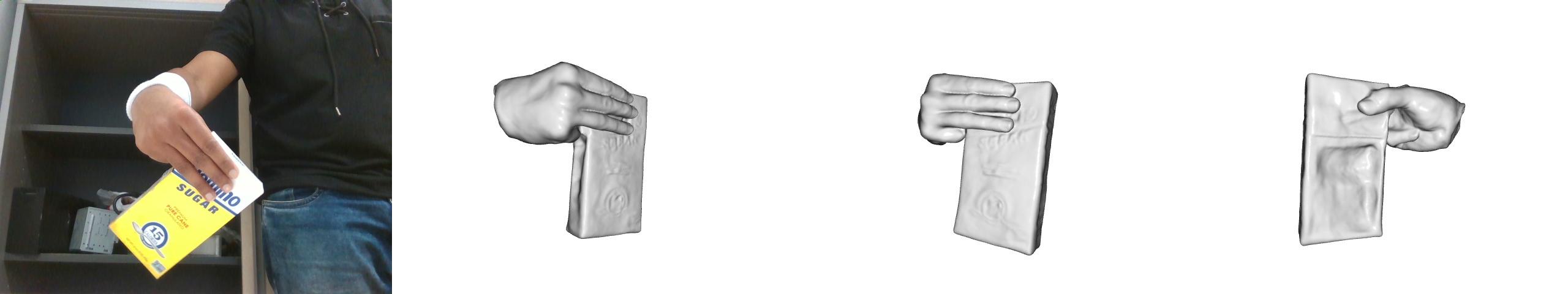}
    \includegraphics[width=\linewidth]{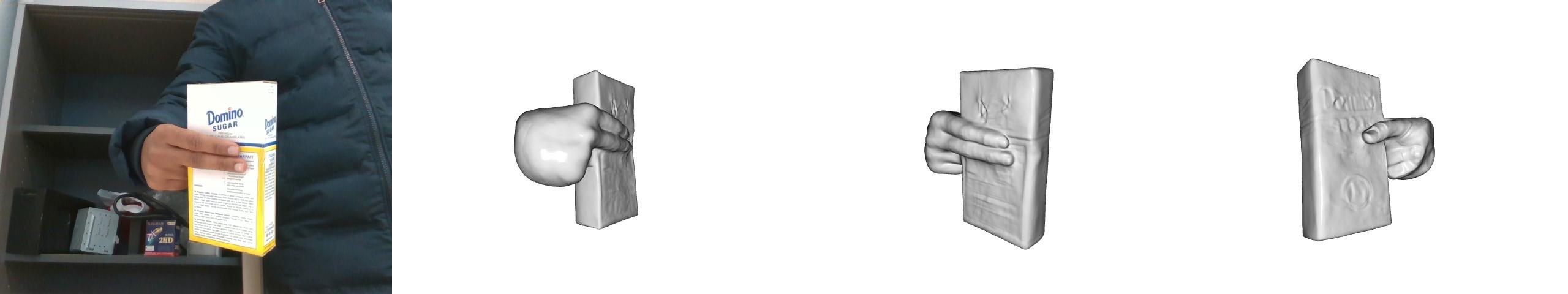}
    \vspace{-0.5cm}
    \caption[Qualitative Hand-object reconstructions on HO3D dataset]{\textbf{Qualitative Hand-Object reconstructions on the HO3D dataset.} The first column is an RGB frame of the input sequence, followed by 3 different views of the reconstructed hand-object shape using HOSt3R, our proposed method.}
    \label{fig:quali_dust3r_res_bat1_ho3d}
\end{figure}

\begin{figure*}[t]
    \centering
    \begin{tabular}{@{}c@{}c@{}}
        \begin{minipage}{0.05\textwidth}
            \rotatebox{90}{\textbf{RGB Frames}}\\[20pt]
            \\
            \rotatebox{90}{\textbf{3D views}}
        \end{minipage} &
        \begin{minipage}{0.9\textwidth}
            \includegraphics[width=\textwidth]{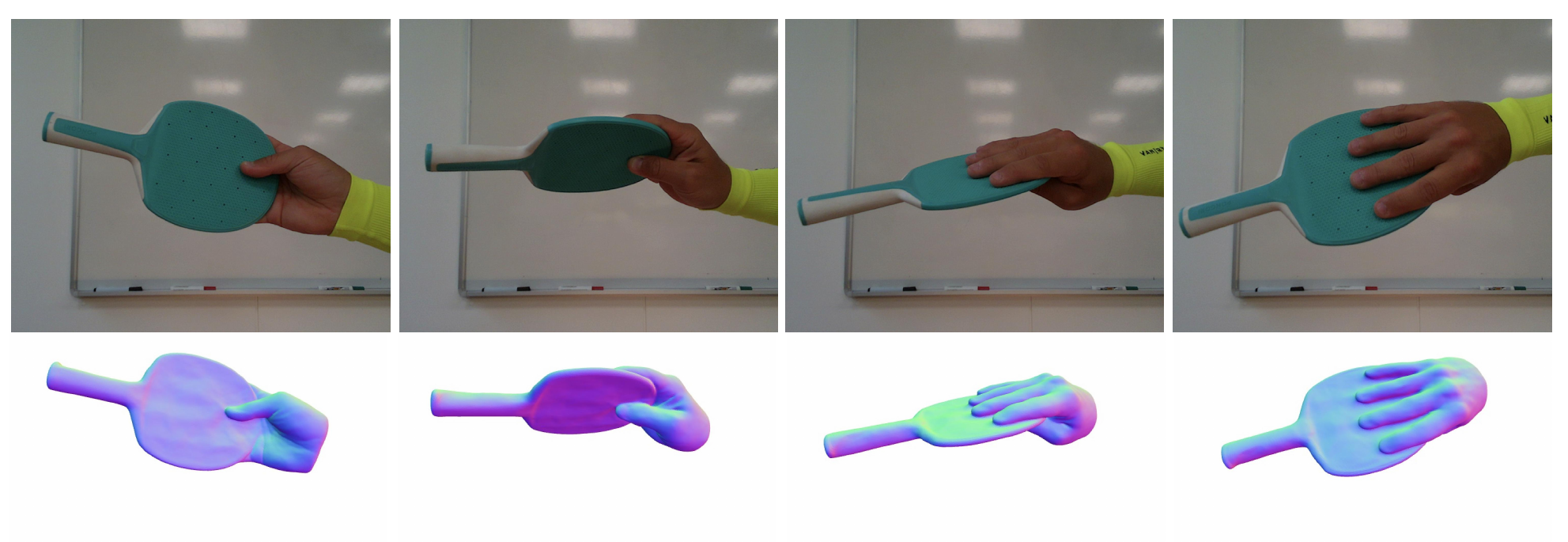}
        \end{minipage} \\[4pt]

        \begin{minipage}{0.05\textwidth}
            \rotatebox{90}{\textbf{RGB Frames}}\\[20pt]
            \\
            \rotatebox{90}{\textbf{3D views}}
        \end{minipage} &
        \begin{minipage}{0.9\textwidth}
            \includegraphics[width=\textwidth]{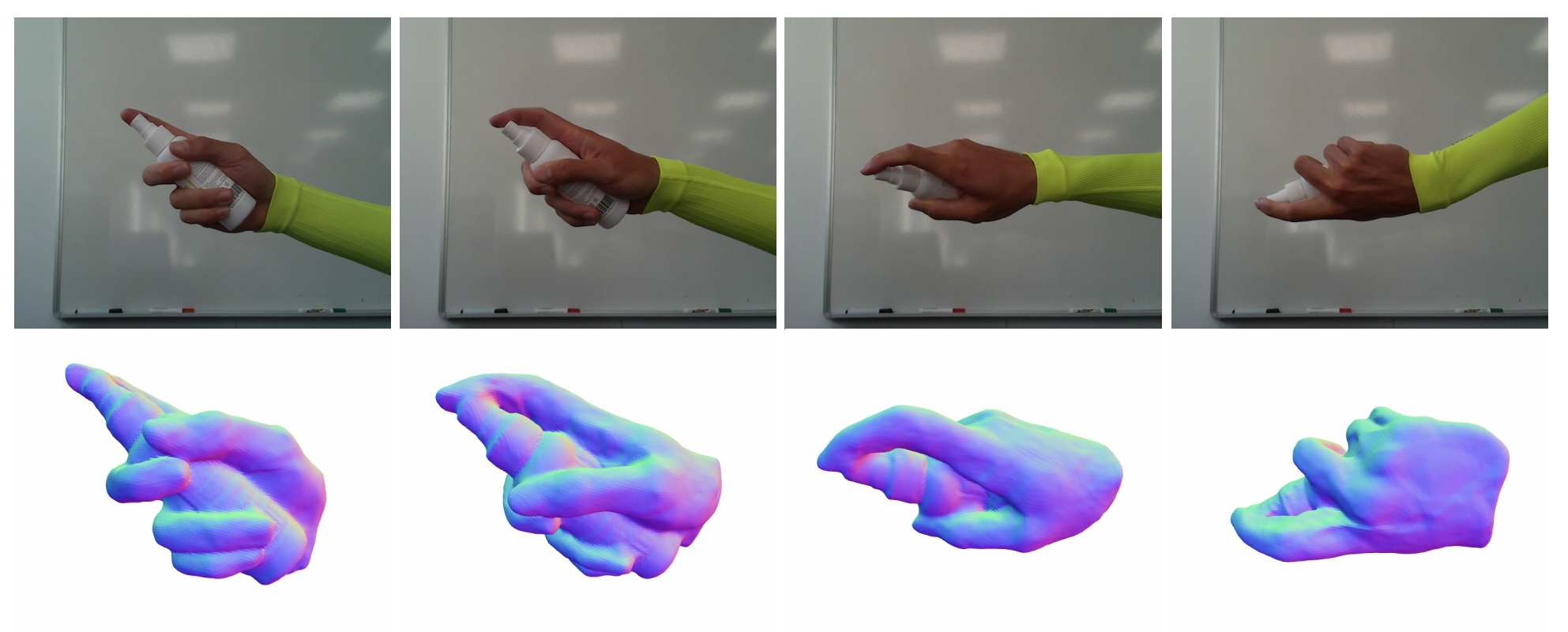}
        \end{minipage}
    \end{tabular}
    \vspace{-0.5cm}
    \caption{\textbf{Keypoint-free hand-object 3D reconstruction}: Given a monocular video sequence of a hand-object motion with an unknown object, our method reconstructs high-fidelity 3D hand-object surfaces. Each row shows one sequence of the SHOWMe dataset: the input image followed by three views of the reconstructed normals. Best viewed in color.}
    \label{fig:teaser}
\end{figure*}

Understanding and reconstructing hand–object interactions in 3D is a key challenge with broad applications in robotics, augmented/virtual reality (AR/VR), and human–computer interaction. Whether enabling natural user interfaces, immersive experiences, or safe object manipulation in collaborative settings, accurate 3D reconstruction of both the hand and the object is essential.
To address this challenge, we propose a two-stage pipeline that involves hand-object transformation estimation and multi-view reconstruction, designed to achieve robust hand-object reconstruction from monocular video or images capturing rigid hand-object motion (see Figure~\ref{fig:quali_dust3r_res_bat1_ho3d}).

Traditional methods for hand-object 3D reconstruction often rely on multi-stage processes, including parametric hand model prediction or keypoint detection, and Structure from Motion (SfM). Although effective in controlled settings, these approaches face significant limitations when applied to complex, real-world scenarios, particularly when dealing with occlusions, dynamic environments, uniform textures and diverse object shapes. Thus, there is a pressing need for more robust and scalable solutions that can handle these challenges without the constraints of template-based or keypoint-dependent techniques. In this work, we address these challenges by proposing a novel approach to hand-object reconstruction without keypoint detection that offers improved scalability and generalization.

Existing hand-object reconstruction methods often assume the availability of known object templates~\citep{hasson_leveraging_2020,hasson_towards_2022,yang2021cpf,kokic_learning_2019,oberweger2019generalized,hampali2021handsformer,ye_whats_2022, jiang2025hand,wu2024reconstructing}, limiting their applicability in in-the-wild scenarios where such templates may not be available. Other approaches~\citep{hasson19_obman,karunratanakul_grasping_2020,chen_alignsdf_2022,chengsdf} do not require known object templates but are trained on datasets containing a limited number of objects, resulting in poor generalization for unseen objects. More recent methods~\citep{ye_whats_2022,ye2023vhoi} attempt to overcome this limitation by learning object shape priors across six object categories and using these priors for hand-object shape reconstruction. However, while these methods improve generalization on novel objects, the reconstructed shapes tend to lack high fidelity.

Several recent studies~\citep{showme,showme2,huang2022hhor,fan2024hold} have demonstrated the potential of a two-stage pipeline for hand-object reconstruction, using keypoint-based 6DoF hand and object tracking followed by multiview reconstruction using implicit neural representations. The quality of 6DoF tracking is critical for achieving a detailed 3D hand-object reconstruction. For example,~\citep{huang2022hhor} uses 3D hand keypoints for 6DoF tracking, while~\citep{showme,fan2024hold} employ poses derived from structure-from-motion pipelines. In contrast, in ~\citep{showme2}, we introduced a robust hand-object transformation estimation technique that performs well on challenging scenarios, such as small, uniformly textured, or occluded objects, but requires fine-tuning when applied to new datasets or varying environments.

In this work, we propose a keypoint-free framework for joint hand–object 3D reconstruction from monocular hand–object motion videos. Our approach, termed HOSt3R, is inspired by recent advances in scene-level reconstruction and is tailored to the challenges of 
hand–object interactions.
HOSt3R is designed to address key limitations of existing methods such as failure in the presence of untextured objects, mutual hand–object occlusion, and variations in background or camera intrinsics, which often hinder keypoint-based pipelines~\citep{showme,showme2,huang2022hhor,fan2024hold}.

We begin by estimating dense 3D pointmaps (\ie, 3D points for every pixel) for pairs of input images. Using these pointmaps, we compute the relative poses between image pairs, and then apply pose averaging across all pairs to recover global hand–object transformations. These transformations are used to initialize a neural implicit model, which jointly optimizes hand–object shape and motion using differentiable volumetric techniques.

Building upon DUSt3R~\citep{dust3r2023}, we adapt its two-stage pointmap estimation and alignment strategy to the more complex setting of hand–object scenes. While DUSt3R achieved breakthrough results in 3D scene reconstruction, its reliance on scene graph optimization introduces high memory requirements, making it impractical for large datasets or long video sequences. We address this limitation by focusing on pairwise point cloud estimation for hand–object pixels, followed by pose averaging, which enables our method to scale efficiently without the memory overhead of full scene graph optimization. By combining the pairwise estimation network with a pose averaging framework, we can scale to longer sequences. Finally, we incorporate the transformations computed in a multi-view reconstruction pipeline to further enhance the accuracy and detail of the recovered hand-object shapes, see examples in~\cref{fig:teaser}.

In summary, our key contributions are as follows:
\begin{enumerate}
    \item We propose a keypoint-free framework for hand-object transformation estimation that is robust to a variety of objects and camera parameter changes.
    \item We integrate the estimated hand-object transformations within a multi-view reconstruction pipeline to achieve template-free hand-object 3D shape reconstruction.
    \item We benchmark the performance of our hand-object reconstruction method on the SHOWMe benchmark.
    \item We also demonstrate the generalization ability of the proposed framework on the HO3D dataset.
\end{enumerate}

\section{Related work}
\label{sec:rw}
\noindent \textbf{Hand-Object reconstruction methods.}  
Hand-object (HO) reconstruction from a single RGB image or monocular video presents significant challenges due to mutual occlusion between the hand and object, complex motion, and  variability in object shape and appearance. The availability of depth information or a known 3D object models can facilitate shape estimation. Early works~\citep{tzionas_capturing_2014,tzionas_capturing_2016,zhang21,ballan2012motion,oikonomidis2011full} used RGB-D or multi-view inputs to simplify the reconstruction process, but recent advances have focused on using monocular RGB images to achieve similar outcomes.  

Hand-object reconstruction approaches generally fall into two categories: parametric model-based techniques and implicit representation-based methods. Parametric model-based approaches~\citep{hasson_learning_2019, romero_embodied_2017, cao_reconstructing_2021, liu_semi-supervised_2021, kokic_learning_2019, vedaldi_html_2020} typically rely on predefined object templates or category-specific models to estimate hand and object poses. Some methods combine  parametric models with implicit representations~\citep{ye_whats_2022, chen_alignsdf_2022} to enhance reconstruction detail. For example, \citep{ye_whats_2022} assumes known 3D templates and employs Signed Distance Functions (SDFs) to reconstruct hand and object shapes with greater fidelity. Similarly, \citep{qu2023novel} builds separate models for hand and object shapes, but this approach requires camera calibration and online training, limiting its use to known object templates.

In contrast, implicit representation-based methods~\citep{karunratanakul_grasping_2020,qu2023novel} represent  shapes as continuous functions, allowing more flexible and expressive reconstructions. For instance, \citep{chen_alignsdf_2022} reconstructs generic hand-held objects without relying on a pre-defined templates, though the method struggles to generalize across diverse object geometries. These methods face limitations in shape generalization, as they often require training on specific object sets~\citep{karunratanakul_grasping_2020}.

Monocular video-based methods have also gained attention for joint hand-object reconstruction. For example, \citep{hasson_leveraging_2020} leverages photometric consistency over time to improve accuracy, while~\citep{hasson_towards_2022} explores optimization-based approaches. \citep{liu_semi-supervised_2021} uses spatial-temporal consistency to select pseudo-labels for self-training. However, a key limitation of these methods is the reliance on object template meshes during inference, which effectively reduces the reconstruction task to a 6-DoF pose estimation problem. In contrast, our method eliminates the need for object templates, enabling the reconstruction of arbitrary hand-object shapes with high precision.

\paragraph{\textbf{Keypoint-based hand-object transformation.}}
Most hand-object reconstruction methods from multiple RGB images~\citep{showme, huang2022hhor, fan2024hold, ye2023vhoi}  follow a two-stage pipeline: (i) hand-object transformation estimation and (ii) shape estimation from the obtained transformations. Any method that uses either detected 2D hand keypoints~\citep{frankmocap,weinzaepfel2020dope,park2022handoccnet} or salient keypoints~\citep{schoenberger2016sfm} on the image to estimate hand-object transformations is considered keypoint-based.

Several works~\citep{huang2022hhor,ye2023vhoi,showme} employ off-the-shelf hand keypoint detectors~\citep{frankmocap,weinzaepfel2020dope} to compute initial hand–object transformations. These approaches perform well when the hand remains visible (typically in sequences involving small objects) but they fail when the hand is occluded by larger objects. Other methods~\citep{showme,fan2024hold} rely on Structure-from-Motion (SfM) pipelines based on detected image keypoints. 

While effective for scenes with textured or feature-rich objects, this strategy fails in the presence of small or uniformly textured objects due to a lack of salient features.  In contrast to these keypoint-based approaches, we propose a method that does not rely on keypoint detection. Instead, we estimate dense 3D pointmaps, where every pixel is associated with a 3D point, and solve a 2D-to-3D matching problem. This enables robust pose estimation even in the absence of distinctive keypoints.

\paragraph{\textbf{Keypoint-free hand-object transformation.}} 
 An emerging direction in hand–object reconstruction is to directly regress camera parameters~\cite{croco, crocov2} as initialization for a global photometric optimization~\cite{showme, showme2}. This approach generally offers better performance and robustness than keypoint-based methods, in scenes with small or uniformly textured objects. 
However, direct pose regression remains challenging and prone to inaccuracies, as it must disentangle complex interactions between camera intrinsics, extrinsics, and the underlying 3D scene structure~\citep{dust3r2023}. 
To address this, we draw inspiration from~\citep{dust3r2023} and 
tackle pose prediction via pointmap regression: 
a simple over-parameterization of these three quantities that enables easy training and inference, offering strong robustness and generalization capabilities. 
\section{Rigid transformation estimation framework}

\begin{figure*}
    \centering
    \includegraphics[width=\textwidth]{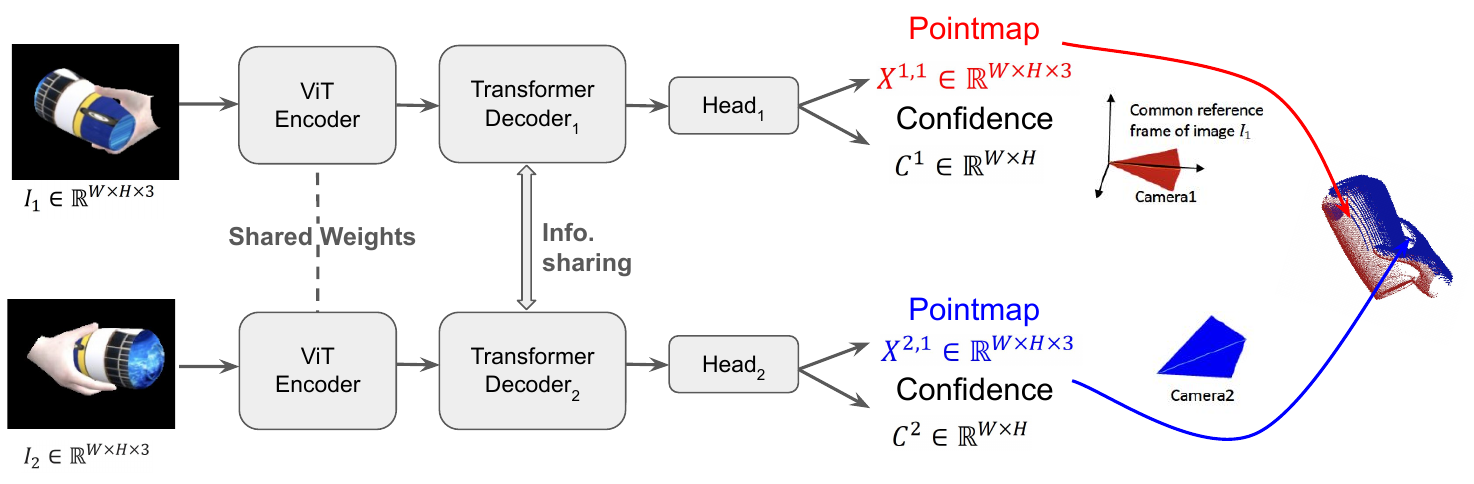}
    \caption[Pairwise pointmaps estimation network]{\textbf{Pairwise pointmaps estimation network.} Given two views of the same hand-object pair \((I_{1}, I_{2})\), the network processes both views through a shared Vision Transformer (ViT) encoder. Each view's features are decoded using distinct decoders that share mutual information. These features are then passed through the `Head' module to predict the pointmap \(X\) and the confidence score map \(C\) for each view. The pointmaps for both views are predicted within the coordinate system of the first view's image, and the network is trained by minimizing the error between the ground truth and the predicted pointmaps.}
    \label{fig:dust3r_net}
\end{figure*}

\subsection{Pairwise pointmap estimation network}
\noindent \textbf{Pointmap.} A pointmap $X \in \mathbb{R}^{W \times H \times 3}$ is a pixelwise prediction where each prediction
represents a 3D point in space. When associated with its corresponding RGB image $I$ of resolution $W \times H$, this pointmap establishes a direct and unique mapping between the image pixels and the 3D points in the scene. Specifically, for each pixel in the image, indexed by coordinates $(i, j)$, there is a corresponding 3D point $X_{i,j}$, such that every pixel $I_{i,j}$ in the image is uniquely linked to a 3D point $X_{i,j}$. Formally, this can be expressed as $I_{i,j} \leftrightarrow X_{i,j}$ for all pixel coordinates $i, j \in \mathbb{N}^{W \times H}$. This mapping assumes that each camera ray intersects with exactly one 3D point, which implies that cases involving translucent or semi-transparent surfaces, where a camera ray might pass through multiple 3D points, are not considered in this scenario. The pointmap, therefore, provides a straightforward representation of the 3D structure of the scene, with each pixel in the image corresponding to a unique location in 3D space. This assumption simplifies the interpretation and processing of the pointmap.

To train the network with strong supervision, we need ground-truth pointmap for every input image. To that end, we consider 
a camera with intrinsic parameters defined by the matrix $K \in \mathbb{R}^{3 \times 3}$. Given this matrix, the pointmap $X$ of the observed scene can be directly computed from the ground-truth depth map $D \in \mathbb{R}^{W \times H}$, where $W$ and $H$ are the width and height of the image, respectively. The relationship between the pointmap $X$ and the depth map $D$ is given by:

\begin{equation}
X_{i,j} = K^{-1} \begin{bmatrix} iD_{i,j} \\ jD_{i,j} \\ D_{i,j} \end{bmatrix}.
\label{eq:gtpointmap}
\end{equation}
Here, $(i, j) \in \mathbb{N}^{W \times H}$ represent the x-y pixel coordinates in the image, and $X$ is expressed in the camera's coordinate frame.

Further, we denote $X^{n,m}$ as the pointmap $X^m$ from camera $n$ expressed in the reference frame of image $m$. This transformation is described by:
\begin{equation}
X^{n,m} = P_m {P_n}^{-1} H(X^n),
\end{equation}
where $P^m, P^n \in \mathbb{R}^{3 \times 4}$ are the world-to-camera pose matrices for views $m$ and $n$, respectively, and $H : (x, y, z) \rightarrow (x, y, z, 1)$ is the homogeneous coordinate mapping. 

\paragraph {\textbf{Pointmap estimation network.}}
We build a pairwise pointmap estimation network $f$ that takes two RGB input images $I_1, I_2 \in \mathbb{R}^{W \times H \times 3}$ as input and produces two corresponding pointmaps $X^{1,1}, X^{2,1} \in \mathbb{R}^{W \times H \times 3}$, along with associated confidence maps $C^{1,1}, C^{2,1} \in \mathbb{R}^{W \times H}$. It is important to note that both pointmaps are expressed within the same reference frame as $I_1$ to ensure consistency across generated outputs. The network $f$'s architecture is inspired by ~\citep{croco,dust3r2023} so that we can benefit from the pretraining of both ~\citep{croco} and ~\citep{dust3r2023}. ~\citep{croco} is trained for cross-view image completion and ~\citep{dust3r2023} is trained for 2D to 3D matching problem. We follow the same architecture for pairwise pointmap estimation as shown in ~\cref{fig:dust3r_net}. Networks $f$ is composed of two symmetrical branches, one of each image comprising an image encoder, decoder, and then the regression module called ``Head'', a sequence of MLP layers. The two input images are first divided into an equal number of patches of size 16$\times$16. Then these patches are processed through a shared ViT encoder~\citep{dosovitskiy2020image} to compute patch embedding (also called as `token') representations $E_{1}$ and $E_{2}$:
\begin{equation}
E_{1} = Encoder(I_{1}), ~~~ E_{2} = Encoder(I_{2}).
\end{equation}

The decoder is a transformer network with cross-attention followed by self-attention. In self-attention, each token attends to every other token of the same view and in cross-attention, a token from the first view attends to every other token from the second view. The decoder module is comprised of blocks of individual decoders and information is always shared between the two image decoders: 

\begin{align}
Z_{i}^{1} &= \text{DecoderBlock}_{i}^{1}\left(Z_{i-1}^{1}, Z_{i-1}^{2}\right), \\
Z_{i}^{2} &= \text{DecoderBlock}_{i}^{2}\left(Z_{i-1}^{2}, Z_{i-1}^{1}\right),
\end{align}
for \(i = 1, \dots, B\) for a decoder with \(B\) blocks and initialized with encoder tokens \(Z_{0}^{1} := E^{1}\) and \(Z_{0}^{2} := E^{2}\). Here, \(\text{DecoderBlock}_{i}^{v}\left(Z^{1}, Z^{2}\right)\) denotes the \(i\)-th block in branch \(v \in \{1, 2\}\), \(Z^{1}\) and \(Z^{2}\) are the input tokens, with \(D^{2}\) the tokens from the other branch. Finally, each image decoder block output is then fed to a ``Head'' MLP network to regress a pointmap and an associated confidence map:
\begin{align}
X^{1,1}, C^{1,1} & = \text{Head}^{1}\left(Z_{0}^{1}, \dots, Z_{B}^{1}\right), \\
X^{2,1}, C^{2,1} & = \text{Head}^{2}\left(Z_{0}^{2}, \dots, Z_{B}^{2}\right).
\end{align}

\paragraph {\textbf{Training Objective.}}
\begin{figure*}[t]
\center
\includegraphics[width=\linewidth]{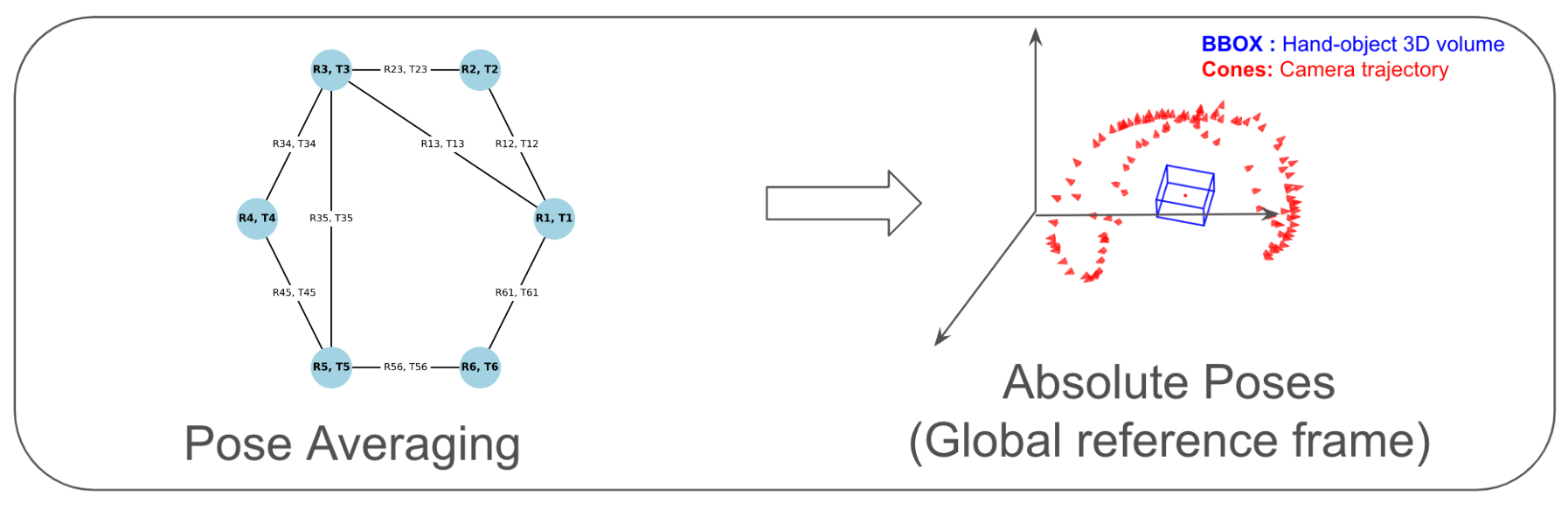}
\caption[Pose averaging]{\textbf{Pose averaging} is a process of estimating absolute poses from a set of relative poses. In the graph, each node $R_k$ represents absolute transformation (rotation and translation) to be optimized, edges represent the measured relative transformation $R_{ij}$. Pose averaging over this graph yields global absolute hand-object transformations of all nodes, plotted as red camera cones in a single coordinate frame.} 
\label{fig:PoseAveraging}
\end{figure*}

The network $f$ is trained using 3D points regression loss with confidence aware terms. 

It builds upon a standard regression loss $\ell_{\text{regr}}$ where we use the Euclidean distance between the ground-truth and predicted pointmaps.
Let us denote the ground-truth pointmaps as $\bar{X}^{1,1}$ and $\bar{X}^{2,1}$, with two corresponding sets of hand-object pixels $\mathcal{D}^1$, $\mathcal{D}^2 = \{i^1, \ldots, i^L\}$ on which the ground-truth is defined. The regression loss for a hand-object mask $M$, pixel $i$ in view $v \in \{1, 2\}$ is defined as below:

\begin{equation}
\ell_{\text{regr}}(v, i) = M_i \cdot | \frac{X^{v,1}_i}{z} - \frac{\bar{X}^{v,1}_i}{\bar{z}} | .
\label{eq:reg_eq}
\end{equation}
\vspace{1mm}

$M_i$ indicates whether the pixel belongs to the hand-object ($M_i=1$) or to the background ($M_i=0$). If $M_i=0$ then pixel $i$ is ignored for the loss calculation. We normalize the predicted and ground-truth pointmaps by scaling factors to handle scale ambiguity between the predictions and ground truth pointmaps. The scaling factors for ground truth and predictions are $z = \text{norm}(X^{1,1}, X^{2,1})$ and $\bar{z} = \text{norm}(\bar{X}^{1,1}, \bar{X}^{2,1})$, respectively. This basically represents pointmap as the distance of all pointmaps from the origin:

\begin{equation}
    \text{norm}(X^1, X^2) = \frac{1}{|\mathcal{D}^1| + |\mathcal{D}^2|} \sum_{v \in \{1,2\}} \sum_{i \in \mathcal{D}^v} \|X^v_i \| .
\end{equation}

To make $\ell_{\text{reg}}$ confidence aware, we enable
the joint prediction of score for each pixel which indicates the confidence of the pointmap prediction. The final loss is the confidence weighted regression loss from ~\cref{eq:reg_eq} for all hand-object pixels:

\begin{equation}
    \ell_{\text{conf}} = \sum_{v \in \{1,2\}} \sum_{i \in \mathcal{D}^v} C^{v,1}_i \ell_{\text{regr}}(v, i) - \alpha \log C^{v,1}_i,
\end{equation}

Here, \(C^{v,1}_i\) represents the confidence score for the \(i\)-th pixel prediction, while \(\alpha\) serves as a hyper-parameter controlling the regularization term. To guarantee that the confidence score remains strictly positive, it is typically defined as \(C^{v,1}_i = 1 + \exp(C^{v,1}_i) > 0\). 

\subsection{Relative pose computation from pointmaps}

The aligned pointmaps from the two views have several useful properties: they are expressed in a common coordinate system, are spatially aligned, and maintain pixel-level correspondence. As a result, the estimated pointmaps can be used to compute the relative pose between the two views. Given two images \(I_1\) and \(I_2\) with corresponding estimated pointmaps \(X^{1,1}\) and \(X^{2,1}\)(in $I_1$ coordinate system), we can compute the relative pose between the two views as follows:
\vspace{1mm}

1. Compute Focal Length from Depth ($X^{1,1}_z$):
   \begin{equation}
   f = \text{estimateFocalLength}(X^{1,1}_z).
   \end{equation}

2. Set Principal Point at the Center:
\vspace{1mm}
   \begin{equation}
   (c_x, c_y) = \left(\frac{W}{2}, \frac{H}{2}\right),
   \end{equation}
\vspace{1mm}   
   where \(W\) and \(H\) represent the width and height of the image, respectively.

3. Solve PnP with RANSAC:
\vspace{1mm}
   To estimate the relative pose between the two views, solve the PnP (Perspective-n-Point) problem using RANSAC:
    \vspace{2mm}
   \begin{equation}
   [R, t] = \text{PnP\_RANSAC}\left( \text{pixel\_coords}(H,W), X^{2,1}, K \right),
   \end{equation}
    \vspace{2mm}
   where:
   \begin{equation}
   K = \begin{pmatrix} 
   f & 0 & c_x \\
   0 & f & c_y \\
   0 & 0 & 1 
   \end{pmatrix}.
   \end{equation}

Here, \(R\) and \(t\) represent the relative rotation and translation between the two images \(I_1\) and \(I_2\).

\subsection{Pose averaging on relative poses}
The relative pose computed using the output of network $f$ is in an arbitrary local coordinate system. However, for 3D hand-object reconstruction from an input hand-object motion sequence, we need to compute global hand-object absolute transformations. To this end, we create a pairwise graph from a set of images ${I_1, I_2, . . . , I_N}$ for a given sequence. We first construct a connectivity graph \( G(V, E) \) where \( N \) images form vertices \( V \) and each edge \( e = (n,m) \in E \) indicates that images \( I_n \) and \( I_m \) share hand-object visual information. To create a graph, we first create all possible image pairs (a fully connected graph) and then use the classifier from ~\citep{showme2} to filter out the invalid image pairs to reduce the fully connected graph to a sparse graph. The idea is to retain the image pair edges that share enough hand-object visual information. The process of estimating rotation and translation for this type of problem is separable~\citep{dellaert2020shonan}, and we explain how to apply this to the camera pose problem as follows.

\paragraph{\textbf{Rotation Averaging.}}
Given a sequence of frames $\{I_1, I_2, ... I_N\}$, we estimate and refine the relative poses, then construct a directed pairwise graph $\mathcal{G}(\mathcal{V}, \mathcal{E})$, where the $N$ frames correspond to the graph's vertices, and each edge $e=(i, j) \in \mathcal{E}$ encodes the relative rotational relationship between frames $I_i$ and $I_j$. To compute the global rotations from this graph, we apply the Shonan rotation averaging method~\citep{dellaert2020shonan,7898474}, which frames the problem as a factor graph. In this graph, each node represents an unknown absolute rotation to be determined, and the edges, or factors, correspond to the previously estimated relative rotations, which are noisy. The objective is to minimize the sum of the Frobenius norms between the predicted and the measured relative rotations:

\vspace{1mm}
\begin{equation}
    \min_{R \in SO(p)^n}  \sum_{(i,j) \in \mathcal{E}} k_{ij} \lVert R_{j} - R_{i}\hat{R}_{ij}  \rVert_F^2,
    \label{eq:sra}
\end{equation}
\vspace{1mm}
from $SO(3)$ to $SO(p)$ where $p > 3$ until the termination, $k_{ij}$ is the concentration parameter for assumed noise model. 

\vspace{1mm}

\begin{table*}[hbt!]
    \centering
    \resizebox{\linewidth}{!}{%
        \begin{tabular}{ll|ccccc}
            \specialrule{.1em}{.05em}{.05em}
            & \multirow{2}{*}{Method} & \multicolumn{5}{c}{Rigid Transformation Error} \\
             &  & Rot Error $\downarrow$  & Trans Error $\downarrow$ & Det. Rate (\%) $\uparrow$ & @(15cm\&15$^{\circ}$)($\%$) $\uparrow$  & @(30cm\&30$^{\circ}$)($\%$) $\uparrow$ \\
            \midrule
            & DOPE ~\citep{weinzaepfel2020dope} + fixed hand pose & 28.9$^{\dagger}$ & 0.23$^{\dagger}$ & 99.0 & 30.2 & 69.1\\
            & DOPE ~\citep{weinzaepfel2020dope} + median filtering & 28.7 & 0.22 & \bf{100.0} & 30.4 & 69.2 \\
            & DOPE ~\citep{weinzaepfel2020dope} + PoseBERT ~\citep{baradel2022posebert} & 28.0 & 0.22 & \bf{100.0} & 19.5 & 58.2\\
            & COLMAP ~\citep{schoenberger2016sfm} & \textbf{15.9}$^{\dagger}$ & \textbf{0.06}$^{\dagger}$ & 78.3 & \textbf{59.5} & 67.2\\
            & SHOWMe ~\citep{showme2} & 20.9 & 0.12 & \textbf{100.0} & 46.0 & 80.0\\ 
            & \textbf{HOSt3R (ours)} & \underline{18.2} & \underline{0.08} & \textbf{100.0} & \underline{50.5} & \textbf{86.3}\\ 
            \specialrule{.1em}{.05em}{.05em} 
        \end{tabular}%
    }
    \vspace{-0.1cm}
\caption[Rigid transformation estimation comparison]{\textbf{Rigid transformation estimation comparison.}
The `Rot. error' is the geodesic distance expressed in degrees with the ground-truth rigid transformation. The `Trans error' is the MSE. $^\dagger$ means the metrics are computed for the frames for which the pose is successfully recovered.}
\vspace{0.2cm}
\label{tab:dust3r_transformations}
\end{table*}

\paragraph{\textbf{Translation Averaging.}}
Translation averaging involves estimating the absolute translations from the previously obtained absolute rotations $\hat{R}_{i=1..N}$ and the relative translations $\hat{t}_{ij}$. To achieve this, we employ the Gaussian Factor Graph optimization framework~\citep{factor_graphs_for_robot_perception}, which minimizes the sum of squared Euclidean distances between the estimated translations $(t_i,t_j)$ and the measured relative translations $t_{ij}$, while also enforcing constraints based on the relative translation measurements:
\begin{equation}
    \min_{{t}_0, t_1, \ldots, t_N} \sum_{(i, j) \in \mathcal{E}} \left\| R_i \cdot \hat{t}_{ij} - (t_j - t_i) \right\|^2.
    \label{eq:tra}
\end{equation}

The optimization problem is subsequently addressed using a linear system solver from GTSAM~\citep{gtsam}, with the anchor factor set by fixing the first pose's position at the origin. An illustration of the pose averaging process is provided in ~\cref{fig:PoseAveraging}.
\section{Experimental results}

In this section, we first describe how the estimated hand–object transformations are integrated into a multi-view stereo (MVS) pipeline to reconstruct the 3D shape of the hand and object. We then detail the datasets used for training and evaluation, and finally present quantitative and qualitative results for both hand–object transformation and 3D shape estimation.

\paragraph{\textbf{Integration with multi-view reconstruction.}}

We estimate pairwise relative poses and perform pose averaging across a hand–object scene graph to obtain global transformations in a common coordinate system. These transformations are then integrated into the multi-view 3D hand–object reconstruction pipeline introduced in~\citep{showme2}. Specifically, we use the estimated rigid transformations as initialization for an implicit neural representation method, which reconstructs both the surface geometry and color of the hand–object scene. During reconstruction, we also refine the transformations through joint optimization to correct for any initial inaccuracies. For each sequence, we sample 60 evenly spaced frames, \eg selecting every 15th frame in a 900-frame video, to perform the 3D reconstruction.

\paragraph{\textbf{Datasets.}}
We initialize our pairwise pointmap estimation network with the pre-trained weights of DUSt3R~\citep{dust3r2023}, originally trained on indoor scene datasets. We then fine-tune the network on a synthetically generated dataset. For synthetic data generation, we adapt the pipeline from ObMan~\citep{hasson19_obman}, which produces parametric hand (MANO) grasp poses for approximately 2.7K everyday object models across 8 object categories. We modify this pipeline to generate multi-view data with varied camera intrinsics, extrinsics, and hand–object occlusion ratios. The resulting dataset includes multi-view RGB images, camera intrinsics and extrinsics, and depth maps. Pointmaps are derived from depth maps and intrinsics using Equation~\ref{eq:gtpointmap}. A depiction of the generated multi-view dataset is provided in the supplementary material.

\paragraph{\textbf{Quantitative results.}}
Following the evaluation protocol of~\citep{showme,showme2} we evaluate both rigid transformation estimation and joint hand-object shape reconstruction.  We adopt the same experimental setup in terms of data splits and the number of frames used for evaluation. We report HOSt3R's rigid transformation errors on the SHOWMe dataset, using the same evaluation table as in~\citep{showme2} (see~\cref{tab:dust3r_transformations}).  Our method outperforms the SHOWMe baseline by a margin of 2.7$^{\circ}$ in rotation error and 4.0cm in translation error. In addition, our method successfully recovers transformations for all sequences, achieving a 100\% detection rate. Furthermore, HOSt3R improves over SHOWMe by 4.5\% in the number of frames with translation error under 15cm and rotation error under 15$^{\circ}$. It also achieves the highest percentage of frames with both translation and rotation errors below 30cm and 30$^{\circ}$, respectively. While slightly underperforming COLMAP in some metrics, our method is significantly more robust, achieving a 100\% detection rate and 86.3\% correct frames under the 30cm / 30$^{\circ}$ threshold.

\begin{table*}
	\centering
	\resizebox{\textwidth}{!}{%
		\begin{tabular}{ll|cccccc}
			\specialrule{.1em}{.05em}{.05em}			
			Rigid & Recon. & Rec. rate &  Acc.$^\dagger$ & Comp.$^\dagger$ & Acc. ratio  & Comp. ratio  & Fscore  \\ 
			Transform & Method &  (\%) $\uparrow$ & (cm) $\downarrow$ &  (cm) $\downarrow$ &  @5mm (\%) $\uparrow$ & @5mm (\%) $\uparrow$ & @5mm (\%) $\uparrow$ \\ 
			\midrule
            \color{red}GT  &  IHOI ~\citep{ye_whats_2022} & 87.3 &  0.79  & 1.34 & 41.7 & 37.8 & 39.3 \\ 
                \midrule
			{\color{red} GT} & \vh~\citep{leroyvh} & 93.7 & 0.42 &	0.65 &	67.3 & 	61.6 & 	63.6 \\ 
			{\color{red} GT} & \briac~\citep{toussaint2022fast} &  95.8 & 0.35	& 0.49		&  75.8 	& 	72.0 	& 	73.5	\\ 
                {\color{red} GT} & \sigg~\citep{huang2022hhor} &  \textbf{100.0} & \bf{0.35}	& \bf{0.32}		&  \bf{81.1} 	& 	\bf{83.8} 	& 	\bf{82.3}	\\ 
		
			\midrule  
                DOPE~\citep{weinzaepfel2020dope} &  \briac~\citep{toussaint2022fast}  & 92.7 & 1.02 &	3.18	&31.7 &	15.7	&20.0 \\ 
		    COLMAP~\citep{schoenberger2016sfm} &  \briac~\citep{toussaint2022fast} & 76.0 & 0.64 & 0.79 & 39.3 & 36.2  & 37.6 \\ 
                COLMAP~\citep{schoenberger2016sfm} &  \sigg~\citep{huang2022hhor} & 72.9 & 0.65 & 0.74 & 40.9 & 42.1  & 41.3 \\ 
                SHOWMe~\citep{showme2} &  \sigg~\citep{huang2022hhor} & \textbf{100.0} & 0.61 & 0.62 & 55.6 & 56.0  & 55.6 \\
                \textbf{HOSt3R (ours)} &  \sigg~\citep{huang2022hhor} & \textbf{100.0} & \textbf{0.58} & \textbf{0.59} & \textbf{56.4} & \textbf{57.0}  & \textbf{56.4} \\
            \specialrule{.1em}{.05em}{.05em} 
		\end{tabular}%
	}
\vspace{-0.1cm}
\caption[Hand-object reconstruction evaluation]{
\textbf{Hand-object reconstruction evaluation} using ground-truth and estimated rigid transformations. $^\dagger$ means that the metrics are obtained by computing on the reconstructed mesh only, the failing ones are not taken into account, making direct comparisons between different methods unfair. DOPE refers to the variant `DOPE + fixed hand pose' from Table~\ref{tab:dust3r_transformations}.
}
\vspace{0.4cm}
\label{tab:recon_dust3r}
\end{table*}

We then evaluate hand–object reconstruction on the SHOWMe benchmark, following the protocol introduced in~\citep{showme,showme2}. In~\cref{tab:recon_dust3r}, we report standard metrics including Accuracy, Completion, and F-score.
HOSt3R achieves the best overall performance among all baselines and is on par with the method proposed in~\citep{showme2}. Qualitative results are shown in~\cref{fig:quali_dust3r_res_bat1} for several SHOWMe sequences; additional results are provided in the supplementary material. The reconstructed hand–object geometry is highly detailed and consistent across a variety of grasps, object shapes, sizes, and textures, demonstrating the robustness and generalization capability of the proposed HOSt3R approach.

\paragraph{\textbf{Generalization.}}

Our proposed method demonstrates strong generalization to unseen hand–object sequences from different datasets. We validate this capability on sequences from the HO3D dataset, which features novel objects, hand shapes, backgrounds, and hand–object motions. We present qualitative results on HO3D sequences that align with our experimental settings, specifically, those with rigid hand–object motion and a comparable number of frames, as shown in~\cref{fig:quali_dust3r_res_bat1_ho3d}. Our method successfully reconstructs detailed hand–object shapes without requiring fine-tuning or access to camera intrinsics. However, some parts of the hand, particularly the fingers, are not fully recovered, as illustrated in the last row and last column of~\cref{fig:quali_dust3r_res_bat1_ho3d}. This limitation is primarily due to insufficient viewpoint coverage of the fingers in the input frames, which remains a common challenge in multi-view reconstruction.

\begin{figure}[t]
    \centering

    \begin{minipage}{\columnwidth}
        \centering
        \begin{tabular}{@{}cccc@{}}
            \makebox[0.22\columnwidth]{\centering \textbf{Input RGB}} &
            \makebox[0.22\columnwidth]{\centering \textbf{View 1}} &
            \makebox[0.22\columnwidth]{\centering \textbf{View 2}} &
            \makebox[0.22\columnwidth]{\centering \textbf{View 3}}
        \end{tabular}
    \end{minipage}
    
    \vspace{1pt}

    \includegraphics[width=0.5\textwidth]{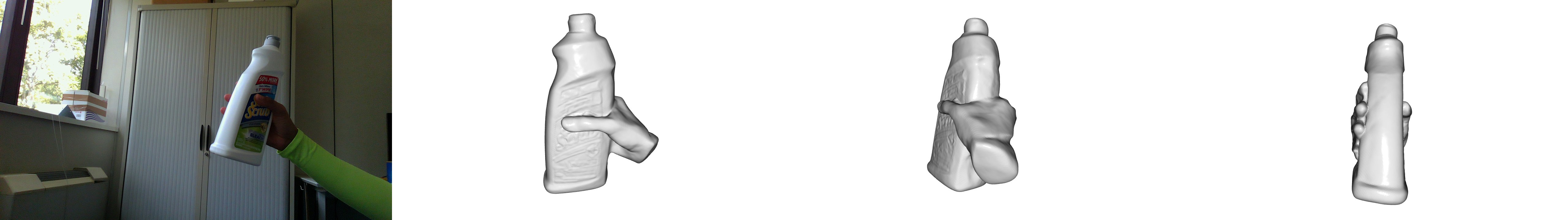}
    \includegraphics[width=0.5\textwidth]{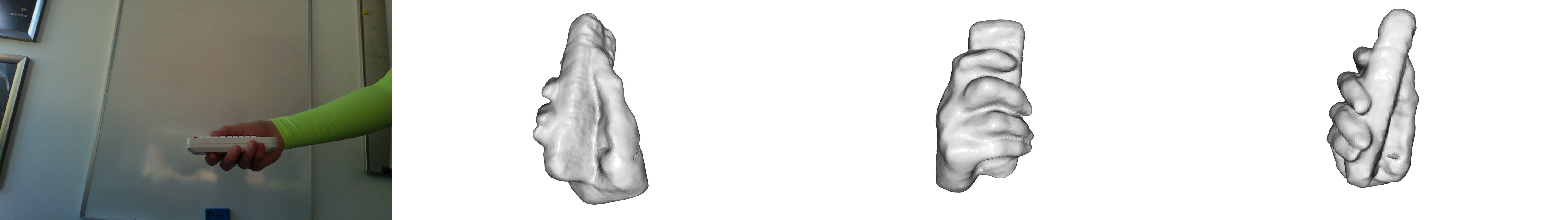}
    \includegraphics[width=0.5\textwidth]{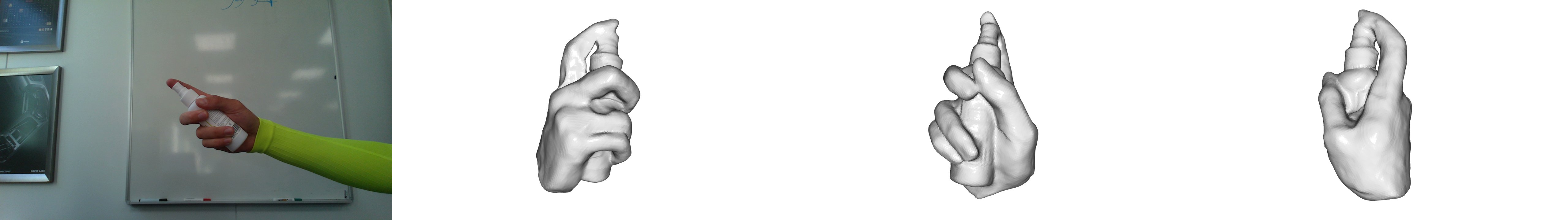}
    \includegraphics[width=0.5\textwidth]{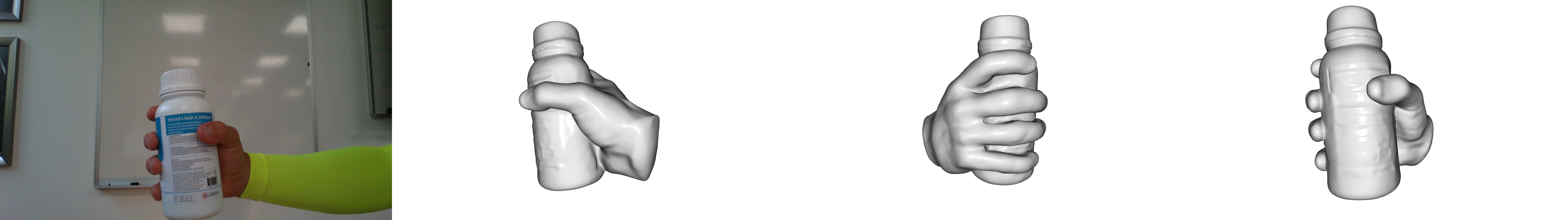}

    \includegraphics[width=0.5\textwidth]{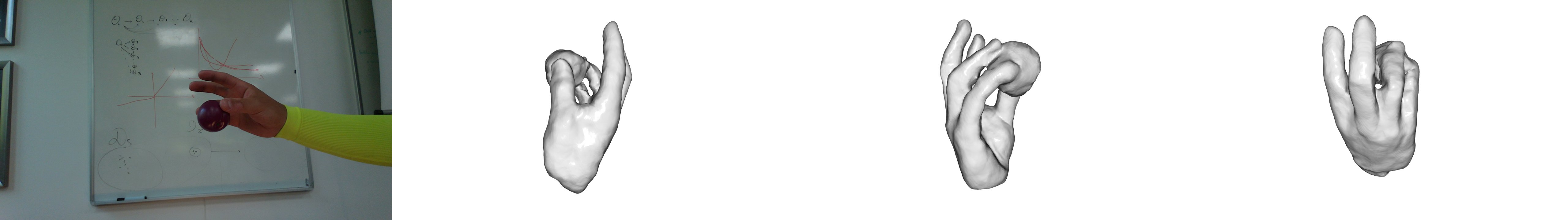}
    \includegraphics[width=0.5\textwidth]{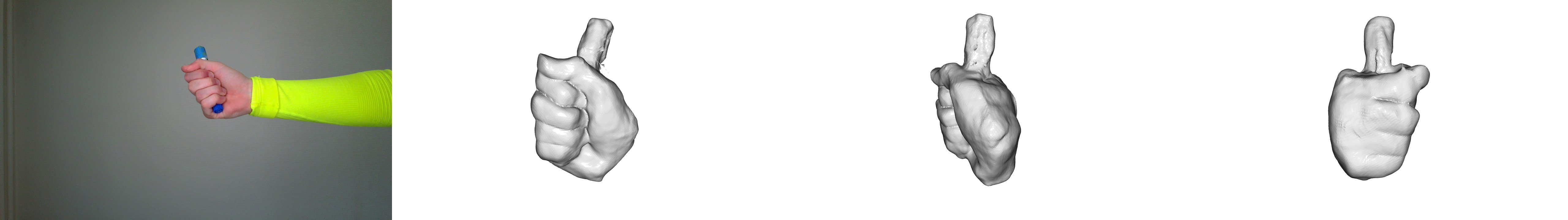}
    \includegraphics[width=0.5\textwidth]{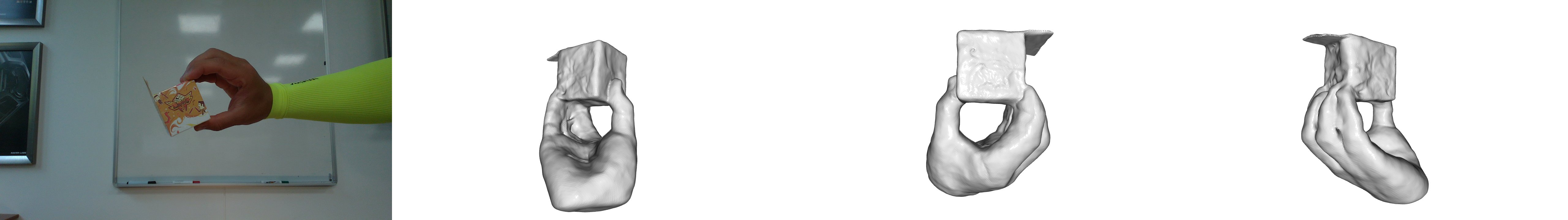}

    \caption{\textbf{Qualitative Hand-object reconstructions on sequences from the SHOWMe dataset.} Each row shows one sequence: the first image is the RGB input, followed by three views of the reconstructed hand-object shape using our method.}
    \label{fig:quali_dust3r_res_bat1}
\end{figure}

\section{Discussion}

We present HOSt3R, a novel method for robust hand–object transformation and 3D reconstruction that overcomes key limitations of prior work. By integrating a DUSt3R-inspired pairwise relative pose estimation network within a pose averaging framework, our approach achieves robustness to camera variations, appearance changes, and occlusions, without relying on keypoint detectors. This design also alleviates memory constraints in large-scale reconstructions.
On the SHOWMe dataset, our method significantly reduces pose errors, achieving a 100\% detection rate and 86.3\% accuracy under the 30cm \& 30$^{\circ}$ error threshold, even in challenging conditions. It generalizes well across object types, grasp styles, and scenes, and offers improved geometric fidelity compared to existing baselines. Qualitative results on the HO3D dataset further demonstrate strong generalization, achieved without fine-tuning or access to camera intrinsics.
While minor limitations remain, particularly in recovering fine finger details under sparse viewpoints, future work could incorporate diffusion-based shape priors~\citep{cheng2024handdiff} to improve reconstruction quality in highly occluded regions.

\clearpage
\setcounter{page}{1}
\maketitlesupplementary

\section{Multi-view Synthetic Dataset}
\label{sec:synthdata}
\begin{figure}[H]
    \centering
    
    \includegraphics[width=0.24\linewidth]{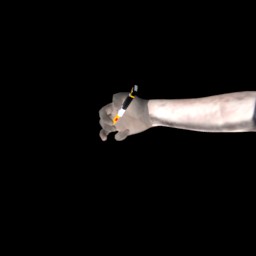}
    \includegraphics[width=0.24\linewidth]{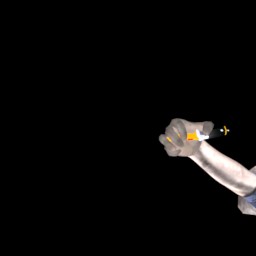}
    \includegraphics[width=0.24\linewidth]{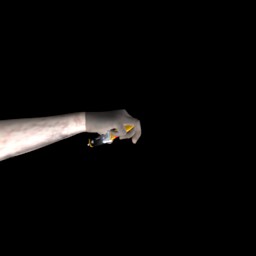}
    \includegraphics[width=0.24\linewidth]{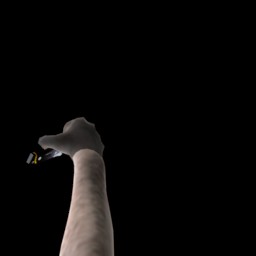}
    \\

    \includegraphics[width=0.24\linewidth]{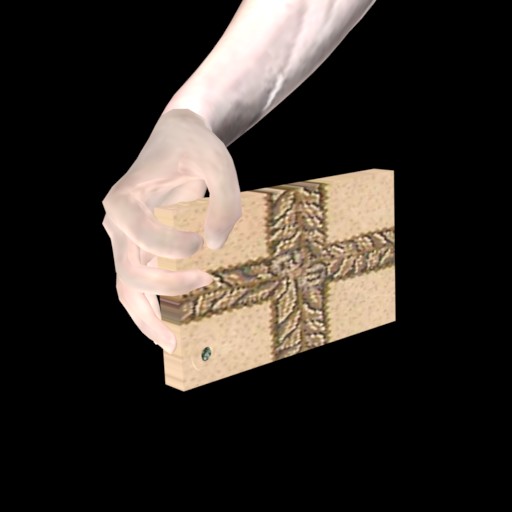}
    \includegraphics[width=0.24\linewidth]{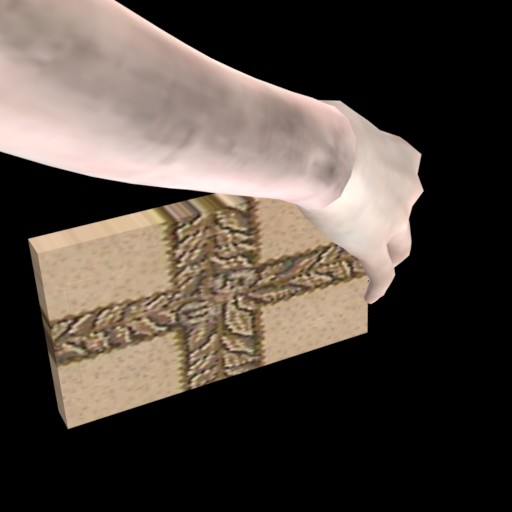}
    \includegraphics[width=0.24\linewidth]{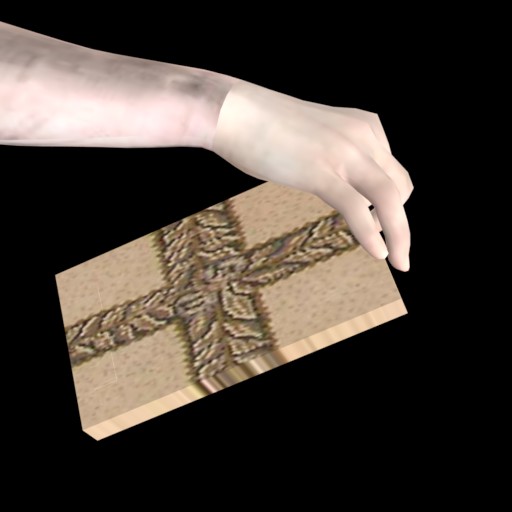}
    \includegraphics[width=0.24\linewidth]{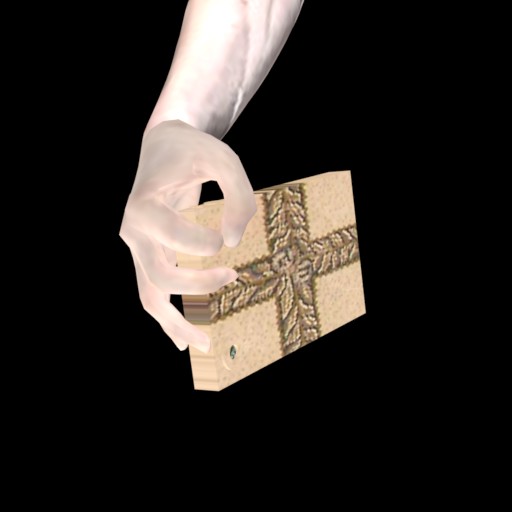}
    \\ 
    \includegraphics[width=0.24\linewidth]{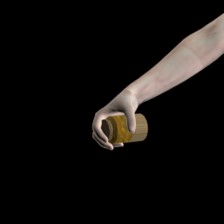}
    \includegraphics[width=0.24\linewidth]{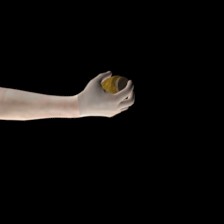}
    \includegraphics[width=0.24\linewidth]{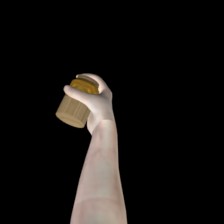}
    \includegraphics[width=0.24\linewidth]{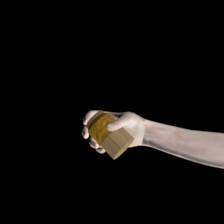}
    \\

    \includegraphics[width=0.24\linewidth]{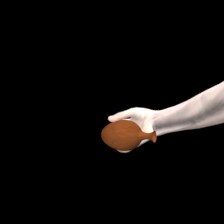}
    \includegraphics[width=0.24\linewidth]{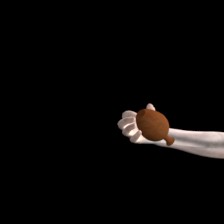}
    \includegraphics[width=0.24\linewidth]{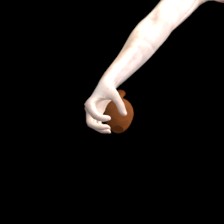}
    \includegraphics[width=0.24\linewidth]{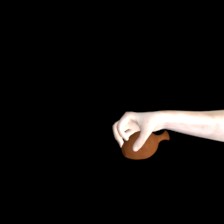}

    \includegraphics[width=0.24\linewidth]{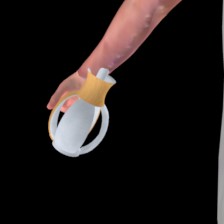}
    \includegraphics[width=0.24\linewidth]{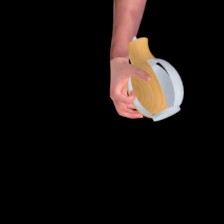}
    \includegraphics[width=0.24\linewidth]{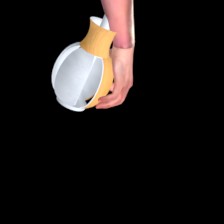}
    \includegraphics[width=0.24\linewidth]{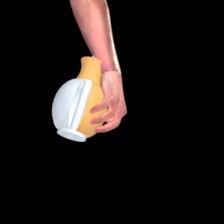}
    
    \caption[Multi-View ObMan synthetic dataset]{\textbf{Example from the Multi-View ObMan synthetic dataset:} Each row represents a single hand-object scene and columns is a different view of the same hand-object. Hand-objects are rendered on a black background, black background is acceptable as pointmap estimation network is trained with only hand-object masked pixels.}
    \label{fig:obman_mv_dset}
\end{figure}

Multi-View ObMan is an extension of the original ObMan~\cite{hasson19_obman} dataset designed to support multi-view hand-object 3D reconstruction tasks. It simulates multiple synchronized camera views of synthetic hand-object interactions, rendered using the MANO~\cite{romero_embodied_2017} hand model and ShapeNet objects~\cite{chang2015shapenet}. Each interaction instance is captured from several virtual viewpoints with consistent lighting and pose, and includes per-view RGB images, depth maps, segmentation masks, 2D keypoints, and full 3D annotations. Camera intrinsics and extrinsics are provided to enable accurate multi-view geometry reasoning. This extension facilitates research in multi-view pose estimation, shape reconstruction, and occlusion-robust modeling of hand-object interactions. A snapshot of RGB frames are depicted in  Figure ~\ref{fig:obman_mv_dset}. 

\begin{figure*}
    \centering
    \includegraphics[width=\textwidth]{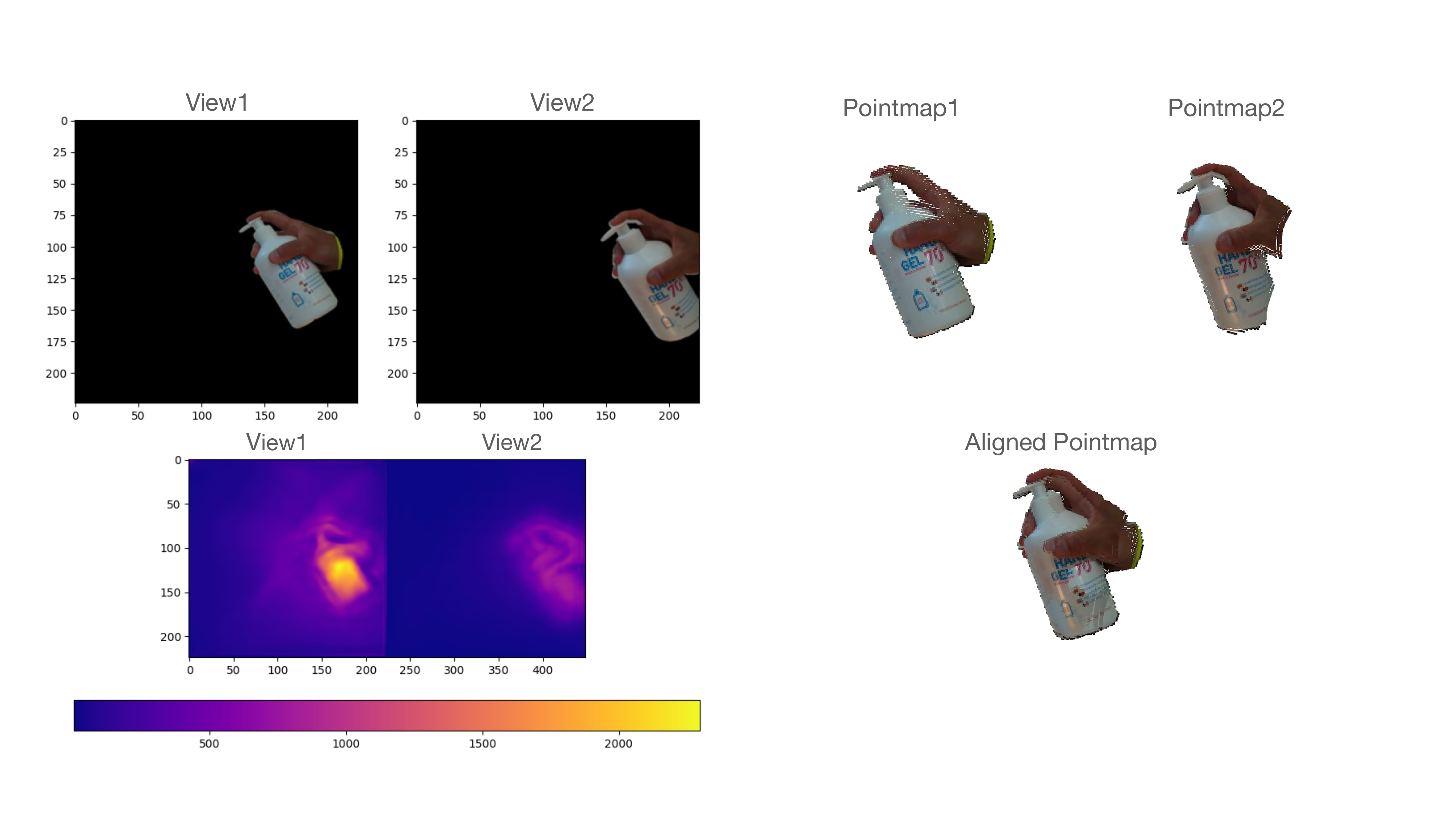}\\
    \textbf{(a)}\\
    \vspace{0.5cm} 
    \includegraphics[width=\textwidth]{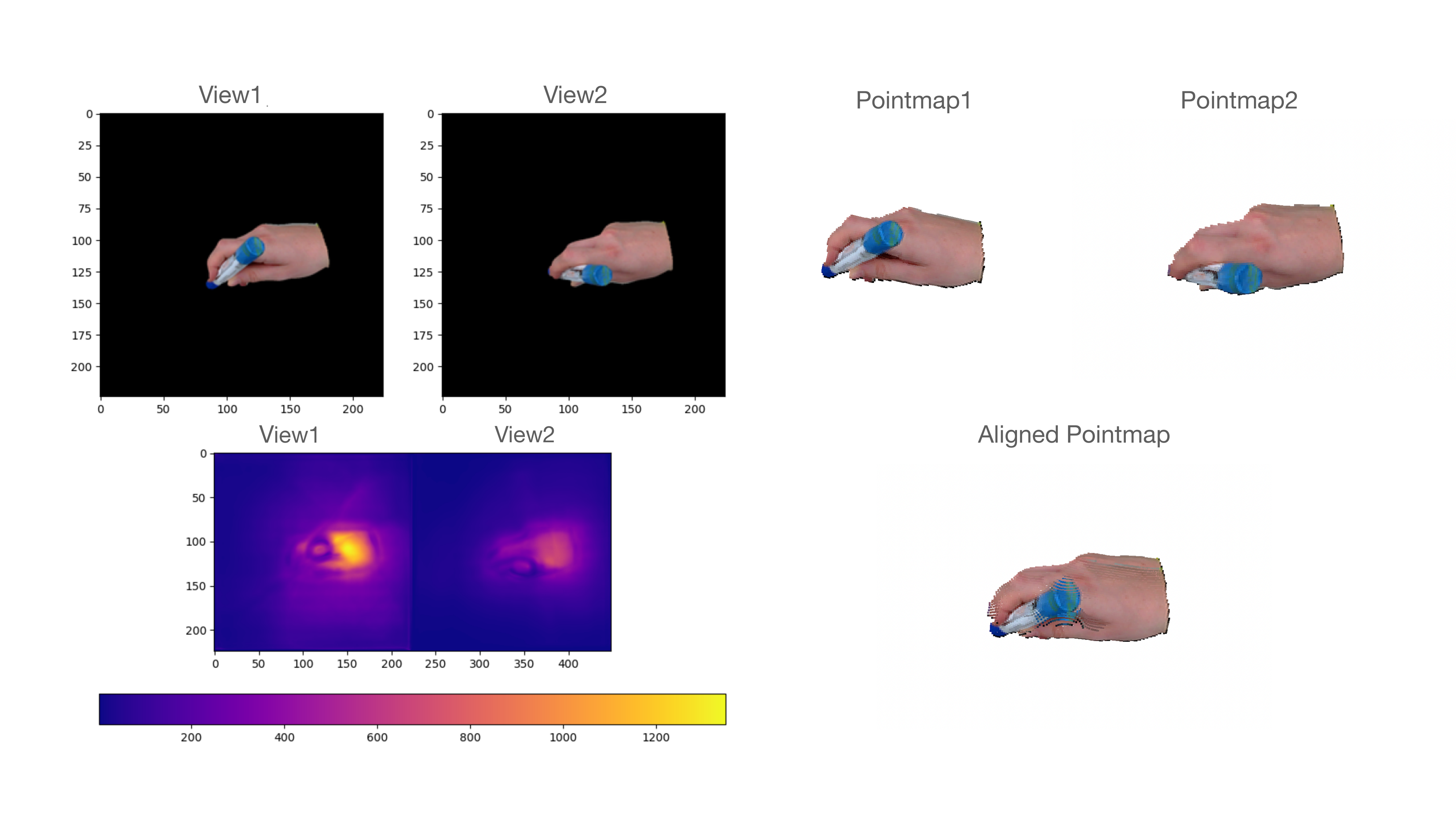}\\
    \textbf{(b)}
    \caption[Pairwise pointmaps estimation results]{\textbf{Pairwise pointmaps estimation results} for two hand-object image pairs (Fig(a) and Fig(b)). In each figure, the top row displays input image pairs and corresponding predicted pointmaps and the bottom row depicts predicted pointmaps for each input image and then predicted aligned pointmaps in the View1 camera coordinate system.}
    \label{fig:pairwise_res}
\end{figure*}

\section{Qualitative results}
\begin{figure}[H]
    \centering

    \begin{minipage}{\columnwidth}
        \centering
        \begin{tabular}{@{}cccc@{}}
            \makebox[0.2\columnwidth]{\centering \textbf{Input RGB}} &
            \makebox[0.2\columnwidth]{\centering \textbf{View 1}} &
            \makebox[0.2\columnwidth]{\centering \textbf{View 2}} &
            \makebox[0.2\columnwidth]{\centering \textbf{View 3}}
        \end{tabular}
    \end{minipage}
    
    \vspace{3pt}

    \includegraphics[width=\columnwidth]{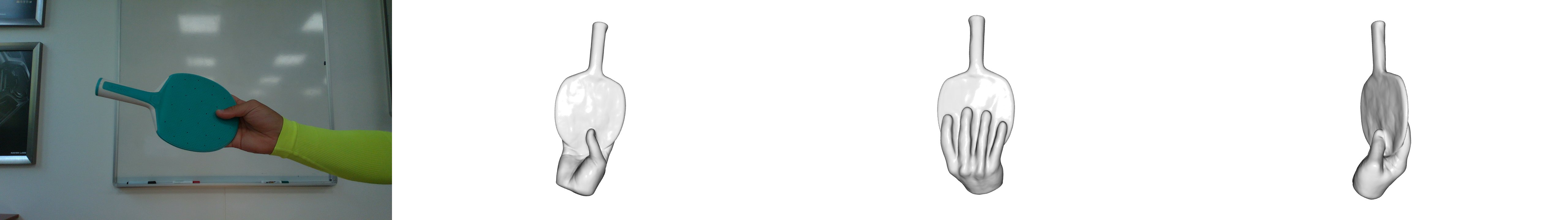}
    \includegraphics[width=\columnwidth]{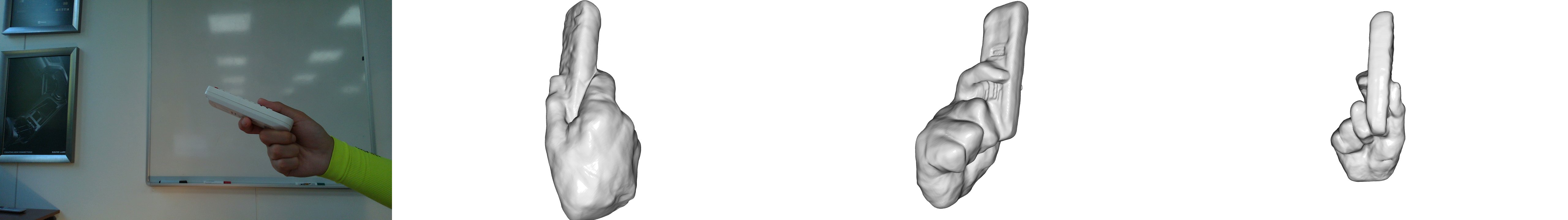}
    \includegraphics[width=\columnwidth]{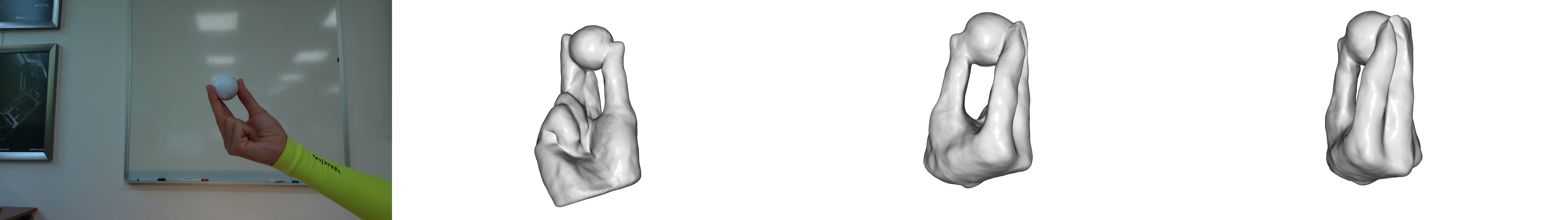}
    \includegraphics[width=\columnwidth]{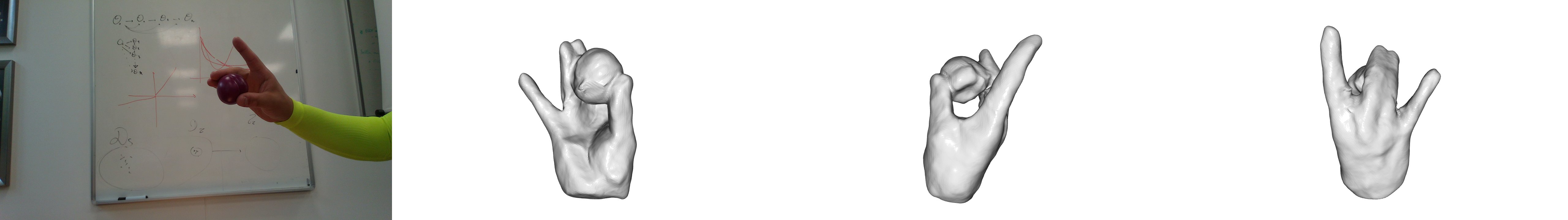}
    \includegraphics[width=\columnwidth]{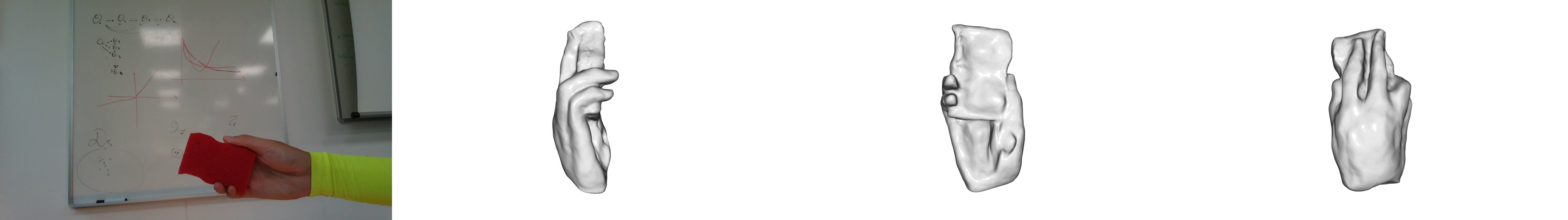}
    \includegraphics[width=\columnwidth]{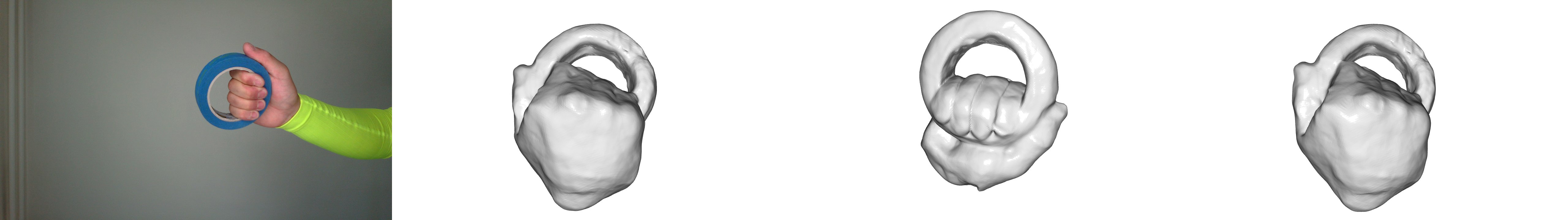}
    \includegraphics[width=\columnwidth]{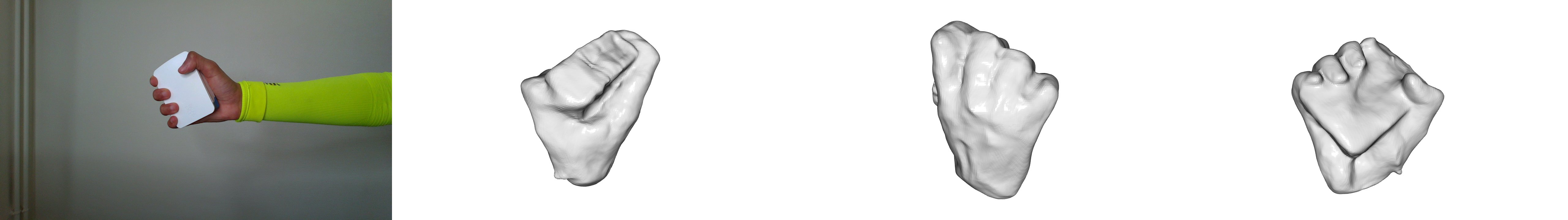}
    \includegraphics[width=\columnwidth]{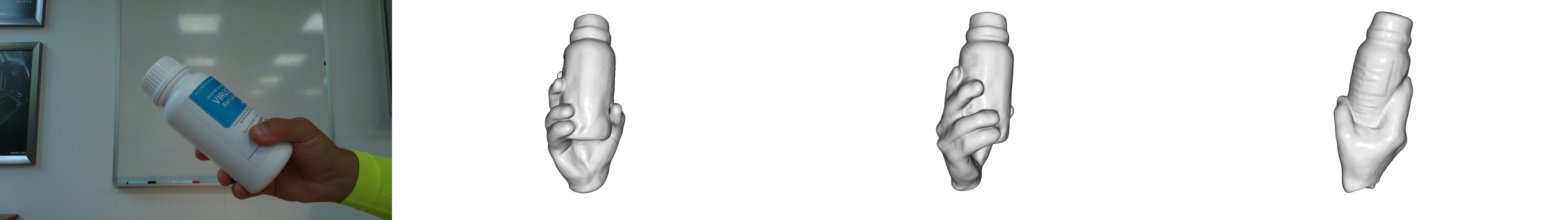}

    \caption{\textbf{Qualitative hand-object reconstructions on sequences from the SHOWMe dataset.} Each row shows one sequence: the first image is the RGB input, followed by three views of the reconstructed hand-object shape using our method.}
    \label{fig:quali_dust3r_res_bat2}
\end{figure}

Figure~\ref{fig:quali_dust3r_res_bat2} provides additional qualitative evaluations on the SHOWMe dataset, focusing on scenarios that pose significant challenges for 3D hand–object reconstruction. These include: (i) thin and elongated structures such as a table tennis racket, which introduce ambiguities in depth estimation due to limited surface area; (ii) small-scale objects like a golf ball, where reduced pixel coverage amplifies sensitivity to noise and calibration errors; and (iii) uniformly textured objects such as a red scrubber, which lack distinctive visual features. For highly specular objects, the accuracy of the estimated pairwise point maps deteriorates, which in turn introduces pose estimation errors and artifacts in the reconstructed 3D geometry (see row 4, column 2). The pairwise pointmap network takes a pair of images and estimates two pointmaps along with corresponding confidence maps (see Figure~\ref{fig:pairwise_res}).

{
    \small
    \bibliographystyle{ieeenat_fullname}
    \bibliography{main}
}
\end{document}